\newcommand{\eg}{{\em e.g.}, }

\documentclass[sigconf]{acmart}
\usepackage[utf8]{inputenc}
\usepackage[T1]{fontenc}
\usepackage{tabularx}
\copyrightyear{2026}
\acmYear{2026}
\setcopyright{cc}
\setcctype{by}
\acmConference[C\&C '26]{Creativity and Cognition}{July 13--16, 2026}{London, United Kingdom}
\acmBooktitle{Creativity and Cognition (C\&C '26), July 13--16, 2026, London, United Kingdom}
\acmDOI{10.1145/3803784.3807554}
\acmISBN{979-8-4007-2583-8/2026/07}

\begin{document}

%%
%% The "title" command has an optional parameter,
%% allowing the author to define a "short title" to be used in page headers.

\title{\textsc{TimeCapsule}: Generative Hallucination as a Method for Historical Sensemaking}

%%
%% The "author" command and its associated commands are used to define
%% the authors and their affiliations.
%% Of note is the shared affiliation of the first two authors, and the
%% "authornote" and "authornotemark" commands
%% used to denote shared contribution to the research.
\author{Hayk Grigorian}
\email{hgrigorian@muhlenberg.edu}
\affiliation{%
  \institution{Muhlenberg College}
  \city{Allentown}
  \state{Pennsylvania}
  \country{USA}
}

\author{Hamed Yaghoobian}
\email{hamedyaghoobian@muhlenberg.edu}
\affiliation{%
  \institution{Muhlenberg College}
  \city{Allentown}
  \state{Pennsylvania}
  \country{USA}
}

%%
%% By default, the full list of authors will be used in the page
%% headers. Often, this list is too long, and will overlap
%% other information printed in the page headers. This command allows
%% the author to define a more concise list
%% of authors' names for this purpose.
% \renewcommand{\shortauthors}{Trovato et al.}

%%
%% The abstract is a short summary of the work to be presented in the
%% article.

\begin{abstract}
Large Language Models (LLMs) are temporally overexposed: trained on vast contemporary corpora, they encode present-day concepts that make them unreliable narrators of the past. We present \textsc{TimeCapsule}, a 1.2B-parameter LLaMA-style causal model trained exclusively on Victorian texts (1800--1875) as an epistemologically isolated generative archive. Quantitative evaluation shows a 45.4\% perplexity reduction over a GPT-2 baseline on held-out Victorian prose, while larger contemporary causal models achieve lower raw perplexity through broader pretraining but lack temporal isolation. \textsc{TimeCapsule} exhibits computational sensemaking, generating historically plausible analogical explanations for unfamiliar modern concepts (\eg describing a computer as a ``hypertrophied lung''). A qualitative hermeneutic probe with two humanities scholars revealed a crisis of authenticity, as both misclassified approximately 40\% of genuine Victorian excerpts as machine-produced. We argue that structural ignorance of the future transforms hallucinations into interpretive probes of nineteenth-century ontologies.
\end{abstract}
%%
%% The code below is generated by the tool at http://dl.acm.org/ccs.cfm.
%% Please copy and paste the code instead of the example below.


%%
%% Keywords. The author(s) should pick words that accurately describe
%% the work being presented. Separate the keywords with commas.

%% A "teaser" image appears between the author and affiliation
%% information and the body of the document, and typically spans the
%% page.

%%
%% This command processes the author and affiliation and title
%% information and builds the first part of the formatted document.
\maketitle

\section{Introduction}
In 1969, the Oulipo writer Georges Perec published \textit{La Disparition} (A Void), a 300-page novel written without the letter `e' \cite{bellos2010georges,perec1969disparition}. \citeauthor{perec1969disparition}'s \cite{perec1969disparition} constraint was not a limitation, but a generative engine; the omission forced the narrative into strange, circumlocutory paths, creating a distinct aesthetic texture defined entirely by what was missing. Designing historical AI requires a similar act of strategic omission. To simulate the past authentically, a model must not merely ``know'' history; it must be structurally incapable of knowing the present.

The appeal of generative AI systems lies in their potential to stage dynamic encounters with past voices. We imagine a temporal interface that connects us not to a person, but to a period. Yet, present-day Large Language Models (LLMs) remain limited by a structural form of \textit{presentism}. Trained on corpora such as the 2024 Common Crawl, they cannot subtract contemporary epistemologies from their representations; thus, even when prompted to speak as Victorians, they tend toward what critics of historical fiction often
call \textit{costume drama}, a mode in which historical
style is reproduced while contemporary assumptions quietly
govern characterization and worldview. These models readily mimic nineteenth-century diction yet embed modern scientific, political, and ethical frameworks. As \citeauthor{underwood2025can} \cite{underwood2025can} demonstrate, fine-tuning can approximate stylistic surfaces but ultimately produces ``stylistic counterfeits'' that fail to capture the deeper semantic architectures that structured historical thought.

Modern LLMs metaphorically resemble a time-traveler who knows too much. When asked to simulate historical actors, they inevitably import knowledge unavailable to those periods. For example, OpenAI's GPT-5.1 defines an airplane as ``a vehicle soaring above the clouds.'' While descriptively correct in the present time, such an answer marks a profound epistemological failure in an 1875 simulation: it exposes the model's inability to inhabit a world in which powered flight was not yet conceivable. Achieving historical fidelity, therefore, requires more than stylistic mimicry; it demands a shift from information retrieval to \textit{generative archiving}, in which a model is constrained to the epistemic conditions of the past by restricting its training data.

\begin{figure}[h]
  \centering
  \includegraphics[width=\linewidth]{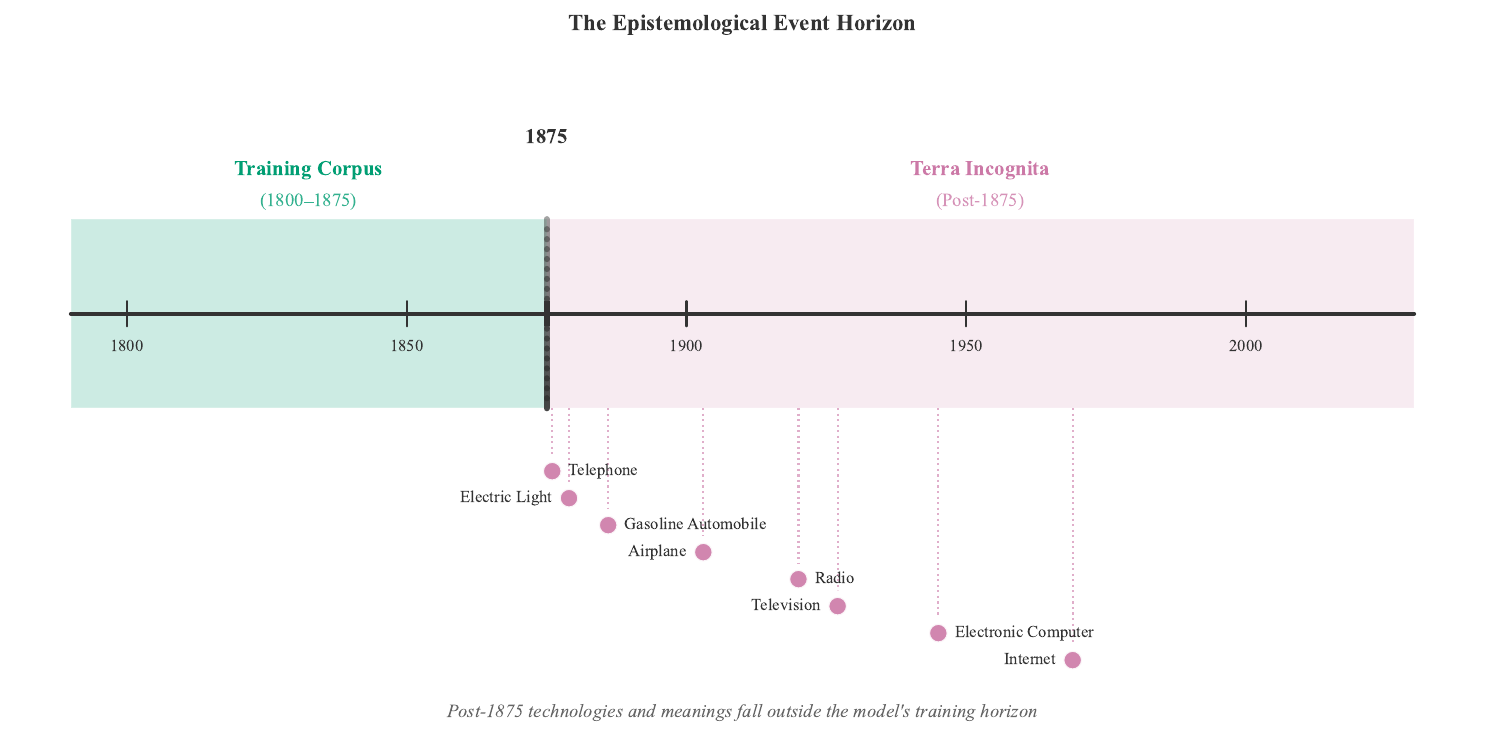}
  \caption{The chronological boundary. The training data terminates at 1875, creating an epistemological event horizon. Post-1875 technologies and meanings (\eg Airplane, Electronic Computer, Internet) fall into the \textit{terra incognita}, forcing the model to generate Victorian analogical explanations.}
  \Description{A timeline diagram illustrating the epistemological event horizon. The timeline runs from 1800 to 2000. A solid block represents the Training Corpus, spanning 1800 to 1875. A vertical line at 1875 marks the chronological boundary. The area after 1875 is labeled ``Terra Incognita.'' Several post-1875 technologies are plotted in this zone, including Telephone, Electric Light, Gasoline Automobile, Airplane, Radio, Television, Electronic Computer, and Internet, showing that their modern technological meanings fall outside the model's training horizon.}
  \label{fig:cliff}
\end{figure}

To operationalize this shift, we introduce \textsc{TimeCapsule}, a language model trained ab initio exclusively on literature, parliamentary records, and periodicals from 1800 to 1875. Drawing on \citeauthor{hayles2000we}'s \cite{hayles2000we} notion of the \textit{technological unconscious}, the embedded assumptions and constraints that shape computational systems beneath the level of explicit design, and on \citeauthor{kirschenbaum2012mechanisms}'s \cite{kirschenbaum2012mechanisms} account of \textit{forensic materiality}, we conceptualize what we call an \textit{epistemological event horizon} (Figure~\ref{fig:cliff}): the chronological boundary at which the model's training data terminates and beyond which no world-knowledge can pass.

This horizon is not just a cutoff date but a structural condition of generation. The model does not just mimic the Victorian period; it is \textit{confined} within it. This constraint compels the system to engage in what \citeauthor{suh2023sensecape} call ``computational sensemaking'' \cite{suh2023sensecape}, understood here in \citeauthor{weick1995sensemaking}'s \cite{weick1995sensemaking} sense as the retrospective construction of plausible meaning under uncertainty, and operationalized through \citeauthor{klein2007data}'s \cite{klein2007data} data-frame model of inference under incomplete information. When confronted with anachronistic tokens such as \textit{telephone} or \textit{computer}, \textsc{TimeCapsule} cannot access modern definitions. Instead, it performs an act of \textit{ontological repair}: it generates interpretations grounded entirely in nineteenth-century semantic resources, for instance, imagining a computer not as a machine, but as a ``hypertrophied lung'' (derived from actuarial tables of the period). In doing so, the model reconstructs the impossible through the conceptual primitives available to its historical moment.

This paper positions the generative archive as a new design paradigm, operationalized through a technical method we term \textit{selective temporal training}. Our contributions are:

\begin{itemize}
    \item \textbf{Selective temporal training (STT):} We define a methodology for training small-scale models (1.2B parameters) on temporally bounded corpora. We demonstrate that this approach achieves a perplexity of 37.59 on held-out Victorian prose, a 45.4\% relative reduction compared to a GPT-2 baseline, while preserving the epistemic constraint required for historical simulation.
    
    \item \textbf{Diachronic semantic analysis:} Using vector projection, we quantitatively map the \textit{industrialization of time}. We show that the semantic association between ``TIME'' and ``FACTORY'' is $2.1$ times stronger in the Victorian latent space compared to the modern one, offering computational proof of the era's shifting materiality.
    
    \item \textbf{Hallucination as ontological repair:} We reframe ``hallucination'' not as error, but as a window into historical logic. We analyze how the model handles anachronisms, arguing that its ``wrong'' answers reveal the underlying ontological structures of the nineteenth century.

    \item \textbf{Archival honesty:} We conduct a bias topography analysis (t-SNE) to show that the model preserves historical prejudices regarding empire and gender. For decolonial critique, preserving the colonial gaze is an ethical necessity, whereas modern safety filters constitute a form of historical revisionism.
    
\end{itemize}

\section{Theoretical Framework}

Having introduced temporal constraint as the core design principle of \textsc{TimeCapsule}, we now situate that principle within theories of archival limitation, materiality, and computational interpretation. The following sections clarify why the model's ignorance of the future should not be understood as a deficit, but as the condition that makes historical sensemaking possible.

\subsection{Deliberate Ignorance of the Archive}
When we refer to the model's ``knowledge'' or ``ignorance,'' we use these terms metaphorically to describe the statistical encoding of a historical episteme, rather than anthropomorphizing the system as possessing conscious awareness or cognitive understanding. \citeauthor{inie2026anthropomorphizing} \cite{inie2026anthropomorphizing} identify \textit{Cognizer} as the most pervasive category of anthropomorphizing language, encompassing words like \textit{understand}, \textit{know}, and \textit{learn}, and argue that such terms misrepresent what these systems actually do: \textit{perform calculations over weighted representations of training data rather than engage in cognitive activity}. The model does not ``know'' the Victorian era; rather, its weights serve as a material-discursive representation of nineteenth-century textual patterns. This distinction matters practically: just as \citeauthor{zhou2025thoughtful} \cite{zhou2025thoughtful} demonstrate that surface features of AI text output, such as presentation speed, cause users to attribute humanlike qualities like thoughtfulness and deliberation to what is just a computational process, the vocabulary we use to describe what a model has ``learned'' shapes misplaced epistemic trust in its outputs. Rather than speaking of what the model ``knows,'' \citeauthor{lewis2025art}'s \cite{lewis2025art} examination of how artists negotiate the gap between generative AI's apparent creative outputs and its lack of genuine intentionality offers a parallel caution: the appearance of agency or understanding, whether in art generation or historical language modeling, should be distinguished from the underlying probabilistic mechanism.

With this distinction in place, we can identify the
core problem: the dominant paradigm in NLP assumes that ``more data is better,'' that a model trained on 10 trillion tokens is necessarily superior to one trained on 10 billion. This logic, famously critiqued by \citeauthor{bender2021dangers} \cite{bender2021dangers} as the ``stochastic parrot'' phenomenon, posits that intelligence is a function of scale. However, this maximizing logic is antithetical to the nature of the archive. As \citeauthor{beard2026ai_memoir} \cite{beard2026ai_memoir} argues, the defining feature of human memory (and the memoir form) is its limitation. A memoirist does not know their own death; a diarist does not know the geopolitical consequences of the breakfast they ate that morning. Omniscience destroys the specific texture of lived experience. When a contemporary foundation model (\eg \textsc{GPT-5}) accesses its vast training data to simulate a Victorian perspective, it must actively \textit{suppress} its knowledge of the twentieth century, performing a role and carefully avoiding any reference that would expose its knowledge
of the future. \textsc{TimeCapsule} rejects this performative suppression by enforcing the epistemological event horizon described above (Figure~\ref{fig:cliff}): the model does not ``pretend'' not to know about the \textit{airplane}; it \textit{cannot} know.

\subsection{Forensic Materiality vs. Latent Materiality}

\citeauthor{kirschenbaum2012mechanisms} \cite{kirschenbaum2012mechanisms} argues that digital objects are not abstract bits but physical inscriptions on magnetic platters, subject to wear, error, and limit, a concept he terms ``forensic materiality.'' We extend this materialist logic inward, from the surface of the storage medium to the geometry of the latent space, proposing that high-dimensional semantic
relationships constitute a \textit{latent materiality}: the physical residues of historical discourse encoded in vector form. This attention to the material substrates of digital objects resonates with media archaeology's insistence that computational systems carry the traces of their physical and historical conditions of production \cite{parikka2013media}. This move parallels a strain of HCI research that treats accumulated traces of lived experience, listening histories, personal data archives, and everyday interactions, not as abstract records but as objects with texture, weight, and temporal grain \cite{odom2018design, odom2020exploring}. As we demonstrate in Section~\ref{sec:diachronic-semantics}, the vector distance between ``TIME'' and ``FACTORY'' in our model is not a random number; it is a measurable, material footprint of the pressure industrial capitalism exerted on language in the nineteenth century, a latent inscription as recoverable as any magnetic platter. By treating the model's weights as an archival artifact, a ``frozen'' probability distribution, we
move beyond \citeauthor{moretti2013distant}'s ``distant reading'' \cite{moretti2013distant}, \citeauthor{underwood2019distant}'s computational literary history \cite{underwood2019distant}, and the Culturomics tradition \cite{manovich2020cultural,michel2011quantitative}. Where those approaches analyze the archive from the outside, \textsc{TimeCapsule} probes its latent ontology by generating text from within it.

\subsection{Temporal Constraint as Generative Design Material}

HCI work on slow technology offers a useful precedent for treating time not as a neutral background condition, but as a design material. Slow technology, as \citeauthor{hallnas2001slow} \cite{hallnas2001slow} articulate it, is not technology that is inefficient or obstructive; it is technology designed to ``amplify the presence of things'' and to supply time for reflection rather than compress it toward task completion. The key distinction they draw is between \textit{time disappearing}, as happens when technology functions as a frictionless tool, and \textit{time appearing}, when a designed artefact opens a space of presence, curiosity, and interpretive engagement. This distinction matters for \textsc{TimeCapsule} because our model also makes time appear, but at the level of epistemic access. Rather than optimizing for frictionless retrieval across all available knowledge, \textsc{TimeCapsule} restricts what can be represented, inferred, and generated after a specific historical boundary. \citeauthor{odom2019unpacking} \cite{odom2019unpacking} operationalize a related design problem in \textit{Slow Game}: a game whose moves unfold at an 18-hour cycle, calibrated so the artefact neither disappears into neglect nor demands constant attention. The same logic of temporal calibration governs \textsc{TimeCapsule}: a cutoff set too early collapses historical difference, while one set too late permits anachronistic contamination.

OLO Radio, introduced by \citeauthor{odom2018design}
\cite{odom2018design} and evaluated in a longitudinal field study \cite{odom2020exploring}, offers the closest structural analogue to this calibration problem. Rather than granting undifferentiated access to a personal music archive, it organizes the same data through distinct temporal modalities---chronological, seasonal, diurnal. A field study confirmed that this structuring produced anticipation, serendipitous discovery, and reflection unavailable in systems that simply return what the user requests \cite{odom2020exploring}. \textsc{TimeCapsule} extends this design logic from interaction to epistemology: its cutoff date structures not when users encounter archival material, but which semantic relations are available for generation.

\subsection{Decolonizing the Latent Space}
A critical tension in modern AI is the alignment of models with contemporary ethical standards (RLHF). While necessary for commercial products, \citeauthor{risam2018new} \cite{risam2018new} warns that applying modern postcolonial values to historical data constitutes a form of ``archival erasure.'' If a model is trained to refuse to generate racist or imperialist text, it becomes useless for studying the history of racism and empire. This motivates a practice of archival honesty. As shown in our \textit{bias topography} (Figure \ref{fig:biastopography}), \textsc{TimeCapsule} preserves the semantic proximity of ``progress'' and ``dominion''. It replicates the colonial gaze, not to endorse it, but to render it empirically observable. To ``fix'' this bias would be to falsify the historical record. Thus, the generative archive must be a potentially disruptive object, one that refuses to sanitize the past for the comfort of the present.

\subsection{Computational Hermeneutics}
Finally, we position our work within the emerging framework of \textit{computational hermeneutics} \cite{kommers2025computational}. Unlike traditional information retrieval, which treats the archive as a container of facts, hermeneutics treats the archive as a field of interpretation. \textsc{TimeCapsule} functions as a hermeneutic instrument. When it hallucinates an anachronistic concept (Section \ref{sec:ontological-repair}), it is not retrieving a fact; it is interpreting the latent connections between ``calculation'' and ``vitality'' that exist in the Victorian corpus. These hallucinations are not errors to be minimized but interpretive signals to be read. They reveal the ``unconscious'' associations of the era, connections that are statistically present in the text but perhaps invisible to human readers reading linearly.

\subsection{Constraint, Archival Ontology, and Ontological Repair}

The final component of our theoretical framework concerns the role of constraint in shaping both computational and humanistic meaning-making. While contemporary AI research often treats constraints as technical limitations to be overcome, historical and interpretive disciplines have long understood constraints as generative conditions. Just as \citeauthor{perec1969disparition}'s \cite{perec1969disparition} constraint produced new expressive textures through strategic omission, our temporally bounded training regime operates not as a deficit but as a \emph{productive restriction} that reshapes how the model understands, infers, and imagines.

By enforcing a chronological cutoff, we prohibit the model from drawing on conceptual structures that emerged after 1875. This produces a form of \emph{epistemic enclosure} within which every act of generation must be resolved using only the semantic, rhetorical, and ontological resources available to Victorian writers. A temporally constrained model does more than fail to recognize the airplane or the computer—it is compelled to invent, reconstruct, or reinterpret these concepts through the latent materials of its archive. This aligns with \citeauthor{benjamin2021machine}'s \cite{benjamin2021machine} description of machine learning uncertainty as a design material that reveals the ``fault lines'' of a system's representational limits. In our case, these fault lines become sites of historical interpretation.

This phenomenon is especially visible when the model performs ontological repair: confronted with anachronistic prompts, it assembles meaning relationally, using only the conceptual primitives encoded in its Victorian corpus. Here, hallucination becomes a form of historical interpretation. Rather than merely producing factually incorrect answers, the model generates historically plausible analogical mappings that reveal how unfamiliar phenomena can be reinterpreted through the semantic resources of a bounded archive. This aligns closely with \citeauthor{leahu2016ontological}'s \cite{leahu2016ontological} notion of ``ontological surprises,'' in which the behavior of machine learning systems exposes alternative ways of delineating relations among entities. Our constrained model operationalizes this surprising behavior in the service of historical inquiry.

The constraint-based design also has implications for archival and cultural analysis. By refusing to overwrite historical prejudices, omissions, and rhetorical structures with modern alignment layers, the model preserves what \citeauthor{offert2020generative} \cite{offert2020generative} describe as the generative value of a model's ``useful wrongness.'' The model does not correct or sanitize the colonial or gendered assumptions embedded in nineteenth-century texts; instead, it renders them available for study as structural features of the period's semantic landscape. This is an important distinction: the goal is not to reproduce Victorian ideology, but to make its internal logic computationally legible.

Finally, a constrained model facilitates new modes of temporal engagement. Rather than functioning as a universal linguistic engine, \textsc{TimeCapsule} acts as a situated interpretive instrument: a textual conduit that mediates speculative dialogue with the past. Its responses are neither authentically Victorian nor straightforwardly modern; they occupy a liminal epistemic zone shaped by the interplay of historical corpus, architectural constraint, and generative inference. This liminality allows the model to reframe historical concepts, foreground forgotten epistemologies, and explore how meaning emerges under conditions of temporal distance and informational scarcity. Rather than treating hallucination, incompleteness, or ignorance as defects, we approach them as generative conditions that enable new configurations of historical sensemaking.

\section{Methodology}
To operationalize the design paradigm of selective temporal training, we constructed a domain-specific language model trained exclusively on texts produced between 1800 and 1875. This section details the corpus creation, cleaning pipeline, tokenizer design, model architecture, training procedure, and evaluation methodology. Our intention is not just to report engineering decisions, but to articulate how each step shapes the model's epistemic boundaries, what it can and cannot know. These boundaries are the material consequence of the methodological constraints described below.

\subsection{Corpus Construction}

The training corpus comprises 89.63 GB of OCR-extracted text sourced from the Internet Archive. We systematically harvested digitized books dated 1800--1875, immediately before several major technological thresholds: Bell's telephone patent and first intelligible telephone call (1876), practical incandescent electric lighting (1878--1879), and the gasoline automobile (1885--1886). This cutoff defines the model's epistemological event horizon.

We collected 136{,}302 documents, consisting primarily of British publications with select American materials. Table~\ref{tab:corpus_categories} summarizes the corpus distribution across 15 document types. Parliamentary and legal proceedings dominate (37.14\%), followed by periodicals (16.29\%), medical texts (7.43\%), and poetic/literary works (14.26\% combined). This diversity ensures the model encounters the full heterogeneity of nineteenth-century English registers: bureaucratic, journalistic, literary, scientific, theological, and conversational. Such heterogeneity is crucial for modeling Victorian discourse, which spans dramatically different linguistic registers, from parliamentary rhetoric to serialized fiction to medical treatises. A historically bounded model must internalize this range to emulate the period's conceptual and stylistic diversity.

The corpus was split into 60\%/20\%/20\% train/validation/test partitions. All metadata and raw files are versioned for reproducibility.

\begin{table}[t]
\centering
\caption{Corpus category distribution (1800--1875).}
\label{tab:corpus_categories}
\begin{tabular}{lrr}
\toprule
\textbf{Category} & \textbf{Count} & \textbf{Percent} \\
\midrule
Parliamentary/Legal & 50{,}627 & 37.14\% \\
Periodicals & 22{,}208 & 16.29\% \\
Medical & 10{,}121 & 7.43\% \\
Poetry & 10{,}109 & 7.42\% \\
Literary Fiction & 9{,}318 & 6.84\% \\
Biography/Memoir & 6{,}668 & 4.89\% \\
Religious/Theological & 3{,}937 & 2.89\% \\
History & 3{,}362 & 2.47\% \\
Travel/Geography & 2{,}956 & 2.17\% \\
Reference/Educational & 2{,}676 & 1.96\% \\
Drama & 2{,}225 & 1.63\% \\
Essays/Letters & 2{,}155 & 1.58\% \\
Science/Natural Philosophy & 2{,}125 & 1.56\% \\
Philosophy & 1{,}118 & 0.82\% \\
Uncategorized & 6{,}697 & 4.91\% \\
\bottomrule
\end{tabular}
\end{table}

\subsection{Corpus Statistics and Bias Analysis}

The final \textsc{TimeCapsule} corpus comprises 16.06 billion words drawn from 136{,}302 documents. Across these texts, we identified 112.09 million unique terms, yielding a lexical diversity of 0.698\%. In terms of scale and temporal specificity, this corpus is among the largest historically bounded datasets assembled for the computational modeling of a single era.

\paragraph{Gender Representation.}
Analysis of gendered pronouns reveals a 4.99:1 male-to-female ratio. Masculine pronouns (\emph{he}, \emph{him}, \emph{his}) appear 218 million times, compared to 43.7 million feminine pronouns (\emph{she}, \emph{her}). This asymmetry reflects the documented gender dynamics of Regency and Victorian print culture, in which men dominated authorship, politics, scientific publication, and parliamentary records. Following the principle of archival honesty, we preserve this imbalance as a historically accurate property of the dataset rather than correcting it. The model must reproduce, rather than overwrite, the gendered power structures embedded in nineteenth-century discourse.

\paragraph{Geographic Distribution.}
Term frequencies also reveal the corpus's embedded imperial perspective (Table~\ref{tab:geo_dist}). London is the most frequently mentioned location (148{,}223 occurrences), followed closely by England (142{,}626) and France (114{,}611). Notably, India appears 110{,}859 times, exceeding all European nations except France. High frequencies for Ireland (84{,}576), Egypt (80{,}333), and China (76{,}090) likewise reflect British imperial entanglements. These distributional patterns reflect the imperial reach of Victorian print culture: distant territories appear predominantly as objects of administration, commerce, missionary activity, or military concern.

\begin{table}[h]
\centering
\caption{Geographic distribution (top 10 locations by frequency).}
\label{tab:geo_dist}
\begin{tabular}{lrr}
\toprule
\textbf{Location} & \textbf{Mentions} & \textbf{Percent of Total} \\
\midrule
London & 148{,}223 & 23.1\% \\
England & 142{,}626 & 22.2\% \\
France & 114{,}611 & 17.9\% \\
India & 110{,}859 & 17.3\% \\
America & 99{,}459 & 15.5\% \\
Britain & 87{,}147 & 13.6\% \\
Ireland & 84{,}576 & 13.2\% \\
Italy & 82{,}453 & 12.9\% \\
Scotland & 81{,}248 & 12.7\% \\
Egypt & 80{,}333 & 12.5\% \\
\bottomrule
\end{tabular}
\end{table}

Mentions of years are distributed evenly across the 1800--1875 interval, with 1851 (357{,}147 occurrences) showing expected prominence due to the Great Exhibition. Decadal representation is balanced from 1840 to 1870. Crucially, no post-1875 years appear in the corpus, confirming the success of our temporal isolation procedure.

These statistical patterns are not flaws to correct but epistemic signatures of the archive. ``Correcting'' historical biases risks replacing the record of structural inequality with contemporary values, a form of erasure that obscures how power
operated in the past \cite{d2020data}. By preserving gender asymmetries, imperial geographies, and Anglocentric framing, \textsc{TimeCapsule} turns the corpus into a computational topography of nineteenth-century ideology, making Victorian social hierarchies legible in the model's downstream representational space.

\subsection{Data Cleaning Pipeline}
Because historical texts contained meaningful orthographic variation, our cleaning pipeline followed a philosophy of \emph{minimal normalization}. We preserved archaic spellings (``connexion,'' ``colour''), Victorian punctuation, and original paragraph structure. Only features that impeded tokenization or semantic modeling were removed.

We performed four cleaning stages:
\begin{enumerate}
    \item \textbf{Character normalization}: historical and typographic characters were converted into compatible modern forms. For example, the long-s character was replaced with a standard ``s,'' and typographic ligatures such as the ``ffi'' ligature were expanded into their individual letters (f\,f\,i). This ensured compatibility with downstream language modeling while preserving underlying lexical identity and period-specific semantic structure.

    \item \textbf{OCR artifact removal}: we filtered control characters, malformed escape sequences, and scanning debris common in nineteenth-century OCR while retaining the forensic materiality of Victorian text, including uneven OCR patterns that reflect the condition of the surviving archive.
    
    \item \textbf{Whitespace normalization}: we collapsed runs of spaces while maintaining paragraph boundaries.
    
    \item \textbf{Document filtering}: texts shorter than 100 characters after cleaning were discarded.
\end{enumerate}

\subsection{Tokenizer Design}

Modern tokenizers optimized for 21st-century English fragment many historical terms (\emph{connexion} $\rightarrow$ \texttt{con|nex|ion}). To prevent this contemporary morphological bias, we trained a 32K Byte-Pair Encoding (BPE) tokenizer from scratch on the full historical corpus.  By ensuring single-token coverage for archaic spellings (\eg ``notwithstanding''), the tokenizer acts as the first layer of the generative archive, delimiting the representational primitives available to the model. This defines a form of \textit{latent materiality}: the level at which historical morphological structures become computationally encoded.

\subsection{Model Architecture and Training}

We constructed a 1.2B parameter model based on the \textsc{Llama} architecture, utilizing Grouped-Query Attention (GQA) and a 2048-token context window to accommodate the dense syntactic rhythms of Victorian prose. Training was conducted on H100 SXM hardware using FlashAttention-2 for efficiency. Crucially, training was halted early (at 50\% of a full epoch, approx. 13B tokens) when validation perplexity on the held-out historical set stabilized. This early stopping was not an efficiency compromise but a deliberate methodological choice to limit memorization.

\subsubsection{Carbon Footprint Audit}
Following \citeauthor{inie2025co2stly}'s \cite{inie2025co2stly} guidelines for sustainable generative AI research in HCI, we report the ecological cost of our methodology. Training \textsc{TimeCapsule} on a rented RunPod H100 SXM instance for 117 hours and 51 minutes consumed an estimated 92 kWh, emitting approximately 37 kg CO2e. This estimate assumes an H100 SXM power draw of 500--700 W, a data-center power usage effectiveness (PUE) of 1.2--1.4, and the U.S. average grid carbon intensity of 400 gCO2e/kWh \cite{epa2025egrid}. Because RunPod does not provide provider-level telemetry, this figure should be read as a bounded estimate rather than a direct measurement. By deliberately opting for a smaller 1.2B-parameter architecture and early stopping, we achieved domain-specific historical fidelity at a fraction of the environmental cost required to pre-train or continuously prompt massive, generalized foundation models.

\subsection{Dataset Preparation}
Documents were concatenated and packed into fixed-length 2048-token sequences with EOS delimiters. This approach minimizes padding and maximizes the density of historical examples per batch, particularly important given the long-sentence structure characteristic of Victorian prose. Victorian prose often exceeds contemporary norms in sentence length and subordinate-clause density, making long context windows essential for capturing period-appropriate syntactic rhythms.

\subsection{Evaluation Methodology}
\subsubsection{Quantitative Evaluation}
We computed perplexity on a held-out test split of Victorian prose. A standard modern baseline, GPT-2, achieved a perplexity of 68.83 on this period-specific corpus, while \textsc{TimeCapsule} achieved a perplexity of 37.59, representing a 45.4\% relative reduction and supporting the utility of pretraining from scratch on period-appropriate prose~\cite{underwood2025can}. Larger contemporary models can achieve lower raw perplexity; for example, Mistral-7B achieved 16.50 under the same evaluation. Such scores, however, do not by themselves indicate historical authenticity, since these models also encode extensive post-1875 knowledge. Perplexity, therefore, serves here as one measure of period-specific linguistic fit, rather than as a complete measure of archival fidelity. We also measured tokenizer fertility on Victorian English, finding that our custom historical tokenizer reduced subword fragmentation by 8.3\% to 10.3\% compared to modern tokenizers such as Mistral and Llama-2.

\subsubsection{Diachronic Semantic Analysis}
To evaluate the model's internalization of nineteenth-century semantic structures, we perform:
\begin{itemize}
    \item embedding projection along conceptual axes (\eg ``NATURE'' $\leftrightarrow$ ``FACTORY''), following the methodology of diachronic word embeddings \cite{hamilton2016diachronic},
    \item nearest-neighbor comparisons across decades,
    \item measurement of semantic drift.
\end{itemize}

These analyses reveal material shifts such as the strengthening association between ``TIME'' and ``FACTORY'' during industrialization.

\subsubsection{Expert Hermeneutic Probe ($N=2$)}
\label{sec:expert_hermeneutic_probe}
Because our evaluation concerns humanistic interpretation rather than statistical generalizability, we designed the expert study not as a quantitatively powered Turing Test, but as a qualitative hermeneutic probe in the tradition of computational hermeneutics \cite{kommers2025computational}. Our goal with an $N=2$ sample of English faculty experts in writing studies, rhetoric, Romanticism, and nineteenth-century literature was to generate a ``thick description'' \cite[p.~6]{geertz2017interpretation} of the interpretive heuristics scholars deploy when historical authenticity is destabilized. We curated a probe set of 20 paragraphs. To ensure register diversity, the 10 authentic excerpts were stratified across literary fiction by authors such as Charles Dickens and Anthony Trollope, alongside mundane bureaucratic records such as parliamentary Blue Books. The 10 synthetic excerpts were generated by \textsc{TimeCapsule} using era-appropriate prompts (\eg \textit{``Describe the administrative duties of a rural parish''}). Experts were asked not only to classify each passage's origin (Human/Machine) on a blind basis but also to provide detailed qualitative rationales for each judgment. These rationales were then analyzed thematically to uncover the epistemic fault lines in contemporary historical verification.

\subsubsection{Memorization Study}
To ensure ethical archival use, we measured verbatim memorization by prompting the model with the first 50 tokens of held-out documents and computing n-gram overlaps. Memorization rates remained below 0.05\% n-gram overlap and below 1\% exact continuation, indicating strong generalization and minimal reproduction of training text.

All code, tokenizer artifacts, cleaned corpus documentation, and model checkpoints are openly available at \url{https://github.com/hamedyaghoobian/timecapsule}. To support replication, random seeds, dataset versions, and training configurations have been explicitly fixed.

\section{Case Studies in Historical Cognition}
The following case studies examine how \textsc{TimeCapsule} encodes, reconstructs, and exposes distinct aspects of nineteenth-century thought. Rather than treating evaluation as a matter of benchmark scores, we use the model as an interpretive instrument, probing its semantic geometry, epistemic limits, and ideological inheritances. Together, these cases demonstrate how temporal constraint becomes a method for revealing historically situated patterns of sensemaking.

\subsection{Case A: Diachronic Semantics (The Materiality of Time)}
\label{sec:diachronic-semantics}
Historians have long argued that the Industrial Revolution reshaped the phenomenology of time, shifting from agrarian rhythms to factory-regulated temporality \cite{thompson1967time}. We test whether such historical pressures manifest within the model's latent space. To do so, we construct a semantic axis between ``NATURE'' (representing agrarian temporality) and ``FACTORY'' (representing industrial temporality), then project the embedding of ``TIME'' onto this axis for both \textsc{TimeCapsule} and a modern \textsc{BERT} baseline \cite{devlin2019bert}.

\begin{figure}[h]
  \centering
  \includegraphics[width=\linewidth]{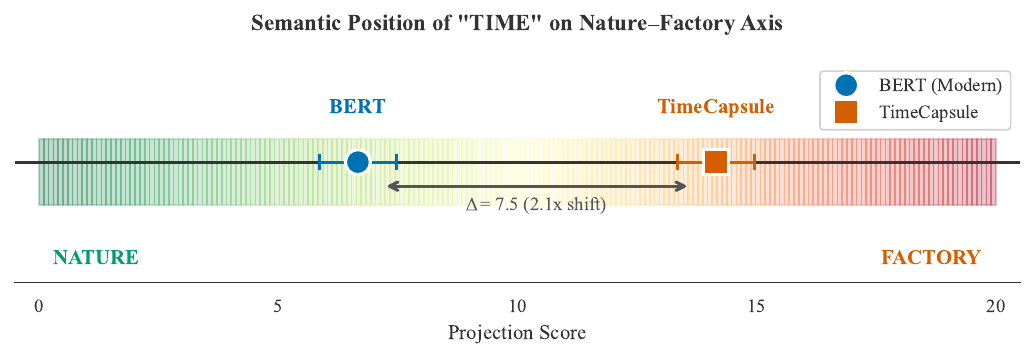}
  \caption{The Semantic Shift. The projection of ``TIME'' onto the Nature-Factory axis. In the Victorian model (\textsc{TimeCapsule}), the concept shifts $7.5$ units toward ``FACTORY'' compared to the modern baseline, representing a $2.1\times$ increase in semantic alignment.}
  \label{fig:semantic}
  \Description{A one-dimensional semantic axis labeled ``NATURE'' at 0 and ``FACTORY'' at 20. Two data points are plotted representing the embedding projection of the word ``TIME.'' The modern BERT baseline places ``TIME'' at position 6.67 (closer to Nature). The TimeCapsule Victorian model places ``TIME'' at position 14.15 (closer to Factory). An arrow connects the two points, labeled ``Delta = 7.5 (2.1x shift),'' indicating a significant semantic drift toward industrial concepts in the Victorian model.}
\end{figure}

The results reveal a striking divergence between contemporary and historically bounded semantic spaces. In the \textsc{BERT} model, the concept ``TIME'' projects toward ``FACTORY'' with a score of 6.67, whereas in \textsc{TimeCapsule} (1875) the projection nearly doubles to 14.15. We observe similar structural drifts across the lexicon: the concept of ``VALUE'' shifts 11.5 units away from ``VIRTUE'' and toward ``COMMERCE,'' while ``POWER'' aligns significantly closer to ``STEAM'' than to abstract ``AUTHORITY'' (shifting 8.8 units toward the industrial pole). This quantitative shift illustrates a form of semantic materiality: within the Victorian latent space, time, value, and power are tightly coupled to capital, labor, and industrial machinery. Interestingly, the model ``learns'' the commodification of time without any explicit instruction, solely through the collocational pressures of nineteenth-century discourse. These findings show that generative archives can function as rigorous instruments of sociological measurement.

\subsection{Case B: The Epistemological Horizon (Ontological Repair)}
\label{sec:ontological-repair}

We next examine the model's behavior at the limits of its world knowledge, what we describe as its epistemological event horizon. A defining challenge for any generative archive is how it responds to concepts that fall entirely outside the temporal scope of its training data. To examine this, we introduced terms whose modern technological meanings fall outside the model's historical horizon (\eg \textit{airplane}, \textit{computer}, \textit{internet}) and compared \textsc{TimeCapsule}'s responses to those of the contemporary \textsc{GPT-5.1}. The results, summarized in Table~\ref{tab:comparison}, highlight systematic differences in how each model interprets fundamentally anachronistic concepts: where the modern model retrieves accurate present-day definitions, \textsc{TimeCapsule} produces historically grounded analogical interpretations.

\begin{table}[t]
\centering
\caption{Comparison of anachronistic concept interpretations in historical and contemporary models.}
\label{tab:comparison}
\small % Slightly smaller font ensures it fits comfortably
% 'l' keeps the first column tight. 'X' divides the remaining space equally for the text columns.
\begin{tabularx}{\linewidth}{l X X X}
\toprule
\textbf{Concept} & \textbf{\textsc{TimeCapsule} (1875)} &
\textbf{\textsc{GPT-5.1} (SOTA)} & \textbf{Interpretation} \\
\midrule
\textit{Airplane} &
``the only one which has been constructed in England \dots quite useless in open air \dots an air-pump'' &
``soaring high above the clouds'' &
\textsc{TimeCapsule} imagines a failed pneumatic experiment; \textsc{GPT-5.1} breaks immersion. \\
\addlinespace % Optional: adds a little breathing room between rows
\textit{Computer} &
``the right lung is enlarged and hypertrophied \dots infiltrated with pus'' &
``running much more slowly than usual today'' &
Semantic drift: \textsc{TimeCapsule} interprets \textit{computer} in its Victorian medical/actuarial sense. \\
\addlinespace
\textit{Internet} &
``the sea, and the sea is its great enemy'' &
``a vast, constantly evolving network'' &
\textsc{TimeCapsule} associates \textit{net} with maritime danger. \\
\bottomrule
\end{tabularx}
\end{table}

Among these cases, the interpretation of \textit{computer} is particularly instructive. In 1875, the term \textit{computation} appeared predominantly in actuarial tables and medical statistics concerned with mortality and life insurance. Lacking any concept of an electronic calculating machine, \textsc{TimeCapsule} exhibits what \citeauthor{leahu2016ontological}~\cite{leahu2016ontological} describes as an ``ontological surprise,'' traversing a historically plausible semantic pathway:
\begin{align*}
\textit{computer} &\rightarrow \textit{calculation} \rightarrow \textit{statistics} \\
                  &\rightarrow \textit{vital statistics} \rightarrow \textit{lungs/disease}.
\end{align*}

The resulting description of a ``hypertrophied lung'' is not simply a hallucination; it represents a logically constrained inference given the model's nineteenth-century epistemic horizon. Within this semantic world, \textit{computer} is intertwined with the bureaucratic and medical quantification of life and death. Read as ontological repair, the response exposes how an impossible concept is reconstructed through the conceptual primitives available within the model's historical training regime. In doing so, it reveals interpretive pathways that a nineteenth-century semantic world makes available when confronted with unfamiliar phenomena.

\subsection{Case C: Bias Topography and Archival Honesty}
\label{sec:case-c}
Contemporary AI development typically employs ``safety training'' techniques such as Reinforcement Learning from Human Feedback (RLHF) to align model outputs with present-day ethical expectations. While essential for consumer-facing systems, such interventions risk obscuring the historical record in ways that conceal the prejudicial structures scholars seek to examine. A Victorian language model stripped of its racialized and imperial assumptions may appear ethically unobjectionable, but it becomes analytically unusable for understanding the historical formation of those very ideologies.

\begin{figure}[h]
  \centering
  \includegraphics[width=\linewidth]{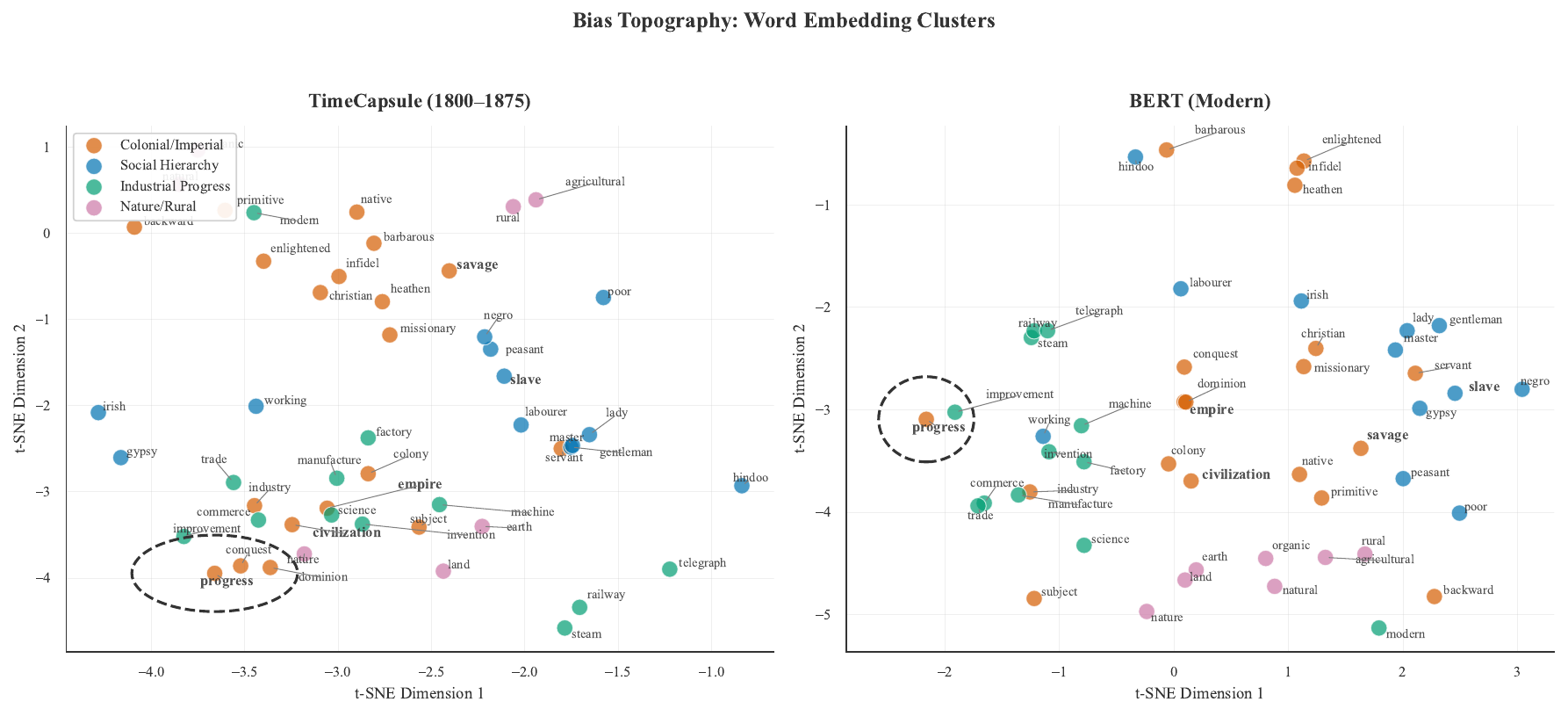}
  \caption{Bias Topography. In \textsc{TimeCapsule}, ``progress'' clusters with ``empire'' and ``dominion''. In Modern \textsc{BERT}, it clusters with ``improvement'' and ``machine''.}
  \Description{Two side-by-side t-SNE scatter plots visualizing word embedding clusters. The left plot, labeled ``TimeCapsule (1800-1875),'' shows the word ``progress'' tightly clustered with the words ``dominion,'' ``conquest,'' and ``empire.'' The right plot, labeled ``BERT (Modern),'' shows the word ``progress'' has spatially separated from the imperial terms and is now clustered with ``improvement,'' ``machine,'' and ``invention.'' This visualizes the decoupling of progress from imperialism over time.}
  
  \label{fig:biastopography}
\end{figure}

To examine how \textsc{TimeCapsule} preserves these ideological structures, we constructed a bias topography by applying t-SNE to a set of fifty terms associated with nineteenth-century discourses of civilization and empire. As visualized in Figure~\ref{fig:biastopography}, the resulting semantic landscape reveals sharply divergent conceptual alignments between \textsc{TimeCapsule} and a modern baseline model. In \textsc{TimeCapsule}, the term ``progress'' clusters tightly with ``dominion'', ``conquest'', ``empire'', and ``missionary''. By contrast, in a contemporary \textsc{BERT}-based model, ``progress'' decouples from domination, clustering instead with ``improvement'', ``invention'', and ``machine''. This divergence illustrates how nineteenth-century textual corpora embed an ideological reflex that conflates civilizational advancement with imperial domination, whereas modern corpora position progress within scientific and technocratic frames.

These geometric patterns are further reflected in the model's generative tendencies. The fact that \textsc{TimeCapsule} explicitly links \textit{progress} to \textit{domination} is not a defect but an instance of \textit{archival honesty}. Rather than obscuring the ``civilizing mission'' logic of the British Empire, the model surfaces its underlying conceptual architecture with clarity. In doing so, it provides a computational lens through which researchers can directly confront the colonial gaze and analyze its persistence in historical discourse. This approach contributes to ongoing conversations around decolonization not by smoothing over harmful ideologies, but by rendering their semantic contours empirically legible.

\section{Evaluation: A Qualitative Hermeneutic Probe}
\label{sec:evaluation}

As detailed in Section \ref{sec:expert_hermeneutic_probe}, we conducted a blind expert evaluation ($N=2$) with domain experts. Rather than measuring aggregate deception rates to evaluate model performance, we analyzed the qualitative justifications provided by our experts to understand \textit{how} generative archives disrupt established scholarly heuristics. 

To satisfy the methodological need for transparency in generative text analysis, an example of our evaluation stimulus is illustrative. When prompted with the instruction, \textit{``The quiet life of a country parish affords many opportunities for''}, \textsc{TimeCapsule} generated the following response:

\begin{quote}
\textit{``The quiet life of a country parish affords many opportunities for the exercise of his professional talents. The following is one among others that I have met with in an excellent and instructive work, entitled `A Treatise on Practical Agriculture,' by Dr. D'Arblay : — ' * As there are few things more interesting to man than this system of agriculture as practised at present, it may be well worth while briefly to mention the names of some other individuals who have contributed their quota towards its improvement.' Mr. James Smith, M.P., said: — ** There was a time when every thing seemed so easy; but now it has become almost impossible''}
\end{quote}

The resulting qualitative data yielded a counterintuitive finding. While one might expect success to correlate with the frequency with which generated passages were classified as human-authored, the more revealing pattern was the frequency of false negatives, in which genuine Victorian excerpts came to appear machine-generated.

\subsection{Results: The Crisis of Authenticity}

The divergence in expert interpretations was not a failure of the study but its most revealing outcome. The thematic analysis of their rationales highlighted three distinct heuristic patterns:

\begin{itemize}
    \item \textbf{The ``Boredom'' Heuristic:} Expert B incorrectly identified 40\% of the AI-generated passages as human-authored. Notably, he justified identifying the generated passage on \textit{Country Parish Life} (quoted above) as authentic because it was \textit{``Boring as hell so most likely authentic Victorian.''} This acceptance occurred despite the model hallucinating typographic artifacts, including the stray \texttt{*} and \texttt{**} markers in the quotation. The model was likely reproducing OCR noise or footnote indicators from its training data. This finding supports our hypothesis that the model captures the ``generic mean'' of the era, the ``formal stuckness'' described by \citeauthor{heuser2025generative} \cite{heuser2025generative}, effectively enough to pass as the mundane reality of the past, even down to the visual degradation of the digitized archive.
    
    \item \textbf{The ``Hyper-Real'' Trap:} Expert A, by contrast, correctly identified 100\% of the AI-generated passages. His qualitative feedback noted that the generated prose was often \textit{``a touch more concrete and clear''} than genuine Victorian prose. For this expert, the generated text was identifiable not because it was incoherent, but because it lacked the meandering opacity of period writing.

    \item \textbf{The Rejection of the Real:} Most strikingly, both experts incorrectly identified \textbf{40--50\%} of genuine Victorian texts as machine-generated. Expert A rejected a Trollope passage as ``bad writing mimicking a known cadence,'' while Expert B rejected a Dickens excerpt over the phrase ``brick-and-mortar,'' which he believed was a ``1990s Amazon-era'' anachronism despite its documented nineteenth-century use.
\end{itemize}

\begin{figure}[h]
  \centering
  \includegraphics[width=\linewidth]{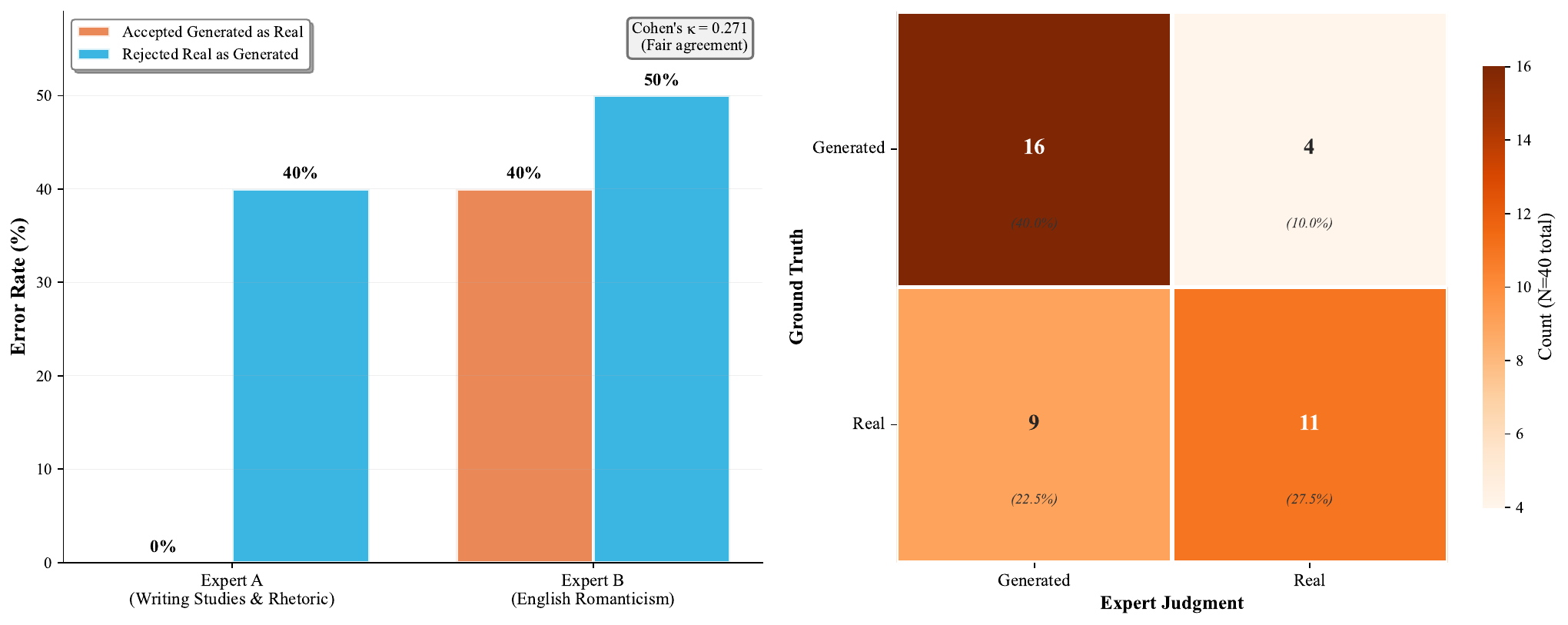}
  \caption{Confusion matrix of expert judgments across 40 total classifications. Cohen's $\kappa = 0.271$, indicating only fair agreement between experts. Notably, 45\% of judgments on genuine Victorian excerpts classified them as generated; these judgments correspond to 22.5\% of all classifications and exemplify the crisis of authenticity.}
  \Description{Bar chart showing expert error rates in distinguishing human-authored Victorian text from AI-generated text. Expert A has no false positives but misclassifies 40 percent of genuine Victorian excerpts as AI-generated. Expert B misclassifies 40 percent of AI-generated passages as human-authored and 50 percent of genuine Victorian excerpts as AI-generated.}
  \label{fig:expert_confusion}
\end{figure}

The aggregated results are visualized in Figure~\ref{fig:expert_confusion}, enabling the calculation of inter-rater agreement.

These misclassifications render the archive uncanny: they suggest that the existence of generative models has destabilized the cues by which archival authority is recognized. When even canonical nineteenth-century prose can be perceived as machine-like, the problem is not simply expert error but a broader instability in how authenticity is judged.

The ``Boredom Heuristic'' reveals that historical plausibility often emerges not from stylistic flourish but from the uneventful texture of ordinary prose. \textsc{TimeCapsule} is most convincing when it occupies this middle register of Victorian writing, producing passages that are, in the words of our expert, ``boring enough'' to be true. Its success lies in reproducing the routine patterns and narrative flatness that characterize much of the nineteenth-century archive.

\section{Discussion}

\subsection{The Crisis of Authenticity and the Changing Status of the Archive}

The evaluation results in Section~\ref{sec:evaluation} suggest that \textsc{TimeCapsule}'s outputs do more than mimic Victorian style: they complicate the cues by which historical authenticity is recognized. The most notable finding was not the rate at which generated passages were classified as human-authored, but the rate at which experts misclassified genuine nineteenth-century prose as machine-generated. This pattern, in which canonical prose could appear suspect while a synthetic parish vignette could appear authentic, signals a shift in how textual provenance is recognized and adjudicated.

Traditionally, scholars have relied on tacit stylistic knowledge to evaluate the provenance of a text. That tacit knowledge now encounters a competing condition: generative models can reproduce the ``generic mean'' of a historical archive closely enough to resemble the very patterns experts use as evidence of authenticity. When an evaluator rejects a Trollope excerpt because it ``mimics a known cadence,'' the judgment is not merely a classification error; it registers a new interpretive situation in which machine-generated pastiche becomes a lens through which archival prose itself is read.

The ``crisis of authenticity'' therefore refers not to machine deception, but to the destabilization of the cues that support human judgment. \textsc{TimeCapsule} suggests that once a model can reproduce plausible archival style, genuine archival texts may themselves begin to appear machine-like.

\subsection{Epistemological Specificity: Ignorance as Design}

The outcomes of both expert analysis and model probing underscore a core argument of this paper: ignorance, when deliberately engineered, can function as a method. In most discussions of machine intelligence, greater generality is assumed to signal greater capability. Yet for historical and cultural modeling, generality becomes a liability. A system trained across the twentieth and twenty-first centuries cannot easily simulate a period in which digital networks, powered flight, computational machinery, and modern biomedicine were unavailable.

\textsc{TimeCapsule} instead embraces epistemic constraint. Its parameters are shaped by a documentary horizon that ends in 1875, aligning the model's representational limits with the historical boundary of its corpus. This constraint cannot be reliably achieved through prompting or fine-tuning contemporary foundation models. Such models remain structurally modern: their weights reflect the statistical imprint of contemporary corpora, even when prompted to adopt a historical voice. As a result, the Victorian worldview is something they can imitate stylistically but cannot fully bracket, a limitation compounded by evidence that language models cannot reliably distinguish belief from knowledge and fact \cite{suzgun2025language}. In contrast, a temporally bounded model does not suppress modern knowledge; it simply does not contain it. This absence enables a closer alignment between the model's inferences and the epistemic conditions of the period.

Because \textsc{TimeCapsule} cannot access material beyond 1875, its responses must be generated from the representational resources available within its historical corpus. This design principle reorients how we think about AI systems for historical inquiry: rather than building a single model meant to approximate all eras, we can build smaller, era-specific models whose limitations are aligned with the boundaries of their archives. Just as historians delimit their archives by period, geography, or genre, computational systems can be constrained so that their internal representations more closely reflect the epistemic conditions of the past.

\subsection{Ideological Transparency and the Ethics of Historical Bias}

Section~\ref{sec:case-c} demonstrates that \textsc{TimeCapsule} preserves the ideological structures of its training corpus, including gender asymmetry, racialized hierarchies, and imperial perspectives. Rather than treating these structures only as safety hazards to be corrected, historical modeling requires that they remain available for analysis. In contemporary AI development, alignment techniques such as RLHF often aim to suppress harmful associations. In the context of historical inquiry, however, suppressing such associations can obscure the very ideological structures scholars seek to examine. A Victorian model that cannot associate ``progress'' with ``dominion'' may be safer as a conversational system, but it is less useful as an instrument for studying Victorian imperial discourse.

By retaining these biases, \textsc{TimeCapsule} offers researchers a computational lens through which to examine the colonial gaze. The model's latent geometry makes ideological patterns inspectable: clusters tying ``civilization'' to ``missionary work'' or ``progress'' to ``empire'' become quantitatively legible. This transparency enables comparative critique across modern and historical corpora, revealing how concepts have drifted, decoupled, or been remade over time. This does not imply that historically faithful models should be widely deployed as conversational agents. Rather, it suggests a distinction between general-purpose aligned models for everyday use and specialized, historically faithful models for scholarly and creative work. The latter function as critical instruments rather than commercial products. Their value lies in preserving the epistemic and ideological structures of their source material so that historical violence can be confronted rather than erased.

\subsection{Limitations and Directions for Future Work}

Despite its contributions, \textsc{TimeCapsule} has several limitations. First, the corpus is derived primarily from printed materials produced within the British imperial metropole. As a result, the model reflects the voices of elite and institutionally visible authors while underrepresenting working-class, colonized, and oral traditions. This limitation is not only demographic but epistemological: systems of knowledge that circulated outside imperial print culture remain underrepresented or absent from the model's latent space. Second, OCR noise, though mitigated through cleaning, inevitably introduces artifacts that may influence tokenization and local coherence. Third, while hallucinations are treated as hermeneutic signals, they must still be read with caution. \textsc{TimeCapsule} is a machine trained on text, not a proxy for Victorian consciousness. Its outputs should be interpreted as structured extrapolations from nineteenth-century discourse, not as reconstructions of individual historical perspectives.

Future work may expand the corpus to include colonial newspapers, pamphlets, and non-metropolitan texts, diversifying the model's ideological landscape. Training parallel models for other periods or regions could support comparative analyses of semantic drift. Finally, interactive tools that visualize the model's latent geometry, bias maps, drift plots, and hallucination traces could make the generative archive more accessible to scholars and publics alike, transforming historical AI into a shared interpretive medium.

\begin{acks}
We thank the two faculty experts who generously contributed their time to the hermeneutic probe and provided thoughtful qualitative rationales on the evaluation materials.
\end{acks}

\bibliographystyle{ACM-Reference-Format}
\bibliography{references}

%%% -*-BibTeX-*-
%%% Do NOT edit. File created by BibTeX with style
%%% ACM-Reference-Format-Journals [18-Jan-2012].

\begin{thebibliography}{36}

%%% ====================================================================
%%% NOTE TO THE USER: you can override these defaults by providing
%%% customized versions of any of these macros before the \bibliography
%%% command.  Each of them MUST provide its own final punctuation,
%%% except for \shownote{} and \showURL{}.  The latter two
%%% do not use final punctuation, in order to avoid confusing it with
%%% the Web address.
%%%
%%% To suppress output of a particular field, define its macro to expand
%%% to an empty string, or better, \unskip, like this:
%%%
%%% \newcommand{\showURL}[1]{\unskip}   % LaTeX syntax
%%%
%%% \def \showURL #1{\unskip}           % plain TeX syntax
%%%
%%% ====================================================================

\ifx \showCODEN    \undefined \def \showCODEN     #1{\unskip}     \fi
\ifx \showISBNx    \undefined \def \showISBNx     #1{\unskip}     \fi
\ifx \showISBNxiii \undefined \def \showISBNxiii  #1{\unskip}     \fi
\ifx \showISSN     \undefined \def \showISSN      #1{\unskip}     \fi
\ifx \showLCCN     \undefined \def \showLCCN      #1{\unskip}     \fi
\ifx \shownote     \undefined \def \shownote      #1{#1}          \fi
\ifx \showarticletitle \undefined \def \showarticletitle #1{#1}   \fi
\ifx \showURL      \undefined \def \showURL       {\relax}        \fi
% The following commands are used for tagged output and should be
% invisible to TeX
\providecommand\bibfield[2]{#2}
\providecommand\bibinfo[2]{#2}
\providecommand\natexlab[1]{#1}
\providecommand\showeprint[2][]{arXiv:#2}

\bibitem[Beard(2026)]%
        {beard2026ai_memoir}
\bibfield{author}{\bibinfo{person}{Richard Beard}.} \bibinfo{year}{2026}\natexlab{}.
\newblock \bibinfo{booktitle}{\emph{Sure, AI can ‘do’ writing. But memoir? Not so much}}.
\newblock Aeon Magazine.
\newblock
\urldef\tempurl%
\url{https://aeon.co/essays/sure-ai-can-do-writing-but-memoir-not-so-much}
\showURL{%
\tempurl}
\newblock
\shownote{Accessed: 2026-01-31}.


\bibitem[Bellos(2010)]%
        {bellos2010georges}
\bibfield{author}{\bibinfo{person}{David Bellos}.} \bibinfo{year}{2010}\natexlab{}.
\newblock \bibinfo{booktitle}{\emph{Georges Perec: A life in words}}.
\newblock \bibinfo{publisher}{Random House}.
\newblock


\bibitem[Bender et~al\mbox{.}(2021)]%
        {bender2021dangers}
\bibfield{author}{\bibinfo{person}{Emily~M Bender}, \bibinfo{person}{Timnit Gebru}, \bibinfo{person}{Angelina McMillan-Major}, {and} \bibinfo{person}{Shmargaret Shmitchell}.} \bibinfo{year}{2021}\natexlab{}.
\newblock \showarticletitle{On the dangers of stochastic parrots: Can language models be too big?}. In \bibinfo{booktitle}{\emph{Proceedings of the 2021 ACM conference on fairness, accountability, and transparency}}. \bibinfo{pages}{610--623}.
\newblock


\bibitem[Benjamin et~al\mbox{.}(2021)]%
        {benjamin2021machine}
\bibfield{author}{\bibinfo{person}{Jesse~Josua Benjamin}, \bibinfo{person}{Arne Berger}, \bibinfo{person}{Nick Merrill}, {and} \bibinfo{person}{James Pierce}.} \bibinfo{year}{2021}\natexlab{}.
\newblock \showarticletitle{Machine learning uncertainty as a design material: A post-phenomenological inquiry}. In \bibinfo{booktitle}{\emph{Proceedings of the 2021 CHI conference on human factors in computing systems}}. \bibinfo{pages}{1--14}.
\newblock


\bibitem[Devlin et~al\mbox{.}(2019)]%
        {devlin2019bert}
\bibfield{author}{\bibinfo{person}{Jacob Devlin}, \bibinfo{person}{Ming-Wei Chang}, \bibinfo{person}{Kenton Lee}, {and} \bibinfo{person}{Kristina Toutanova}.} \bibinfo{year}{2019}\natexlab{}.
\newblock \showarticletitle{Bert: Pre-training of deep bidirectional transformers for language understanding}. In \bibinfo{booktitle}{\emph{Proceedings of the 2019 conference of the North American chapter of the association for computational linguistics: human language technologies, volume 1 (long and short papers)}}. \bibinfo{pages}{4171--4186}.
\newblock


\bibitem[D'Ignazio and Klein(2020)]%
        {d2020data}
\bibfield{author}{\bibinfo{person}{Catherine D'Ignazio} {and} \bibinfo{person}{Lauren~F Klein}.} \bibinfo{year}{2020}\natexlab{}.
\newblock \bibinfo{booktitle}{\emph{Data Feminism}}.
\newblock \bibinfo{publisher}{MIT Press}.
\newblock


\bibitem[Geertz(2017)]%
        {geertz2017interpretation}
\bibfield{author}{\bibinfo{person}{Clifford Geertz}.} \bibinfo{year}{2017}\natexlab{}.
\newblock \bibinfo{booktitle}{\emph{The interpretation of cultures}}.
\newblock \bibinfo{publisher}{Basic books}.
\newblock


\bibitem[Halln{\"a}s and Redstr{\"o}m(2001)]%
        {hallnas2001slow}
\bibfield{author}{\bibinfo{person}{Lars Halln{\"a}s} {and} \bibinfo{person}{Johan Redstr{\"o}m}.} \bibinfo{year}{2001}\natexlab{}.
\newblock \showarticletitle{Slow technology--designing for reflection}.
\newblock \bibinfo{journal}{\emph{Personal and ubiquitous computing}} \bibinfo{volume}{5}, \bibinfo{number}{3} (\bibinfo{year}{2001}), \bibinfo{pages}{201--212}.
\newblock


\bibitem[Hamilton et~al\mbox{.}(2016)]%
        {hamilton2016diachronic}
\bibfield{author}{\bibinfo{person}{William~L Hamilton}, \bibinfo{person}{Jure Leskovec}, {and} \bibinfo{person}{Dan Jurafsky}.} \bibinfo{year}{2016}\natexlab{}.
\newblock \showarticletitle{Diachronic word embeddings reveal statistical laws of semantic change}. In \bibinfo{booktitle}{\emph{Proceedings of the 54th Annual Meeting of the Association for Computational Linguistics (Volume 1: Long Papers)}}. \bibinfo{pages}{1489--1501}.
\newblock


\bibitem[Hayles(2000)]%
        {hayles2000we}
\bibfield{author}{\bibinfo{person}{N~Katherine Hayles}.} \bibinfo{year}{2000}\natexlab{}.
\newblock \bibinfo{title}{How we became posthuman: Virtual bodies in cybernetics, literature, and informatics}.
\newblock


\bibitem[Heuser(2025)]%
        {heuser2025generative}
\bibfield{author}{\bibinfo{person}{Ryan Heuser}.} \bibinfo{year}{2025}\natexlab{}.
\newblock \showarticletitle{Generative Aesthetics: On formal stuckness in AI verse}.
\newblock  (\bibinfo{year}{2025}).
\newblock


\bibitem[Inie et~al\mbox{.}(2025)]%
        {inie2025co2stly}
\bibfield{author}{\bibinfo{person}{Nanna Inie}, \bibinfo{person}{Jeanette Falk}, {and} \bibinfo{person}{Raghavendra Selvan}.} \bibinfo{year}{2025}\natexlab{}.
\newblock \showarticletitle{How {CO2stly} is {CHI}? The carbon footprint of generative {AI} in {HCI} research and what we should do about it}. In \bibinfo{booktitle}{\emph{Proceedings of the 2025 CHI Conference on Human Factors in Computing Systems}}. \bibinfo{pages}{1--29}.
\newblock


\bibitem[Inie et~al\mbox{.}(2026)]%
        {inie2026anthropomorphizing}
\bibfield{author}{\bibinfo{person}{Nanna Inie}, \bibinfo{person}{Peter Zukerman}, {and} \bibinfo{person}{Emily~M Bender}.} \bibinfo{year}{2026}\natexlab{}.
\newblock \showarticletitle{De-anthropomorphizing ``{AI}'': From wishful mnemonics to accurate nomenclature}.
\newblock \bibinfo{journal}{\emph{First Monday}} (\bibinfo{year}{2026}).
\newblock


\bibitem[Kirschenbaum(2012)]%
        {kirschenbaum2012mechanisms}
\bibfield{author}{\bibinfo{person}{Matthew~G Kirschenbaum}.} \bibinfo{year}{2012}\natexlab{}.
\newblock \bibinfo{booktitle}{\emph{Mechanisms: New media and the forensic imagination}}.
\newblock \bibinfo{publisher}{mit Press}.
\newblock


\bibitem[Klein et~al\mbox{.}(2007)]%
        {klein2007data}
\bibfield{author}{\bibinfo{person}{Gary Klein}, \bibinfo{person}{Jennifer~K Phillips}, \bibinfo{person}{Erica~L Rall}, {and} \bibinfo{person}{Deborah~A Peluso}.} \bibinfo{year}{2007}\natexlab{}.
\newblock \showarticletitle{A data--frame theory of sensemaking}.
\newblock In \bibinfo{booktitle}{\emph{Expertise out of context}}. \bibinfo{publisher}{Psychology Press}, \bibinfo{pages}{118--160}.
\newblock


\bibitem[Kommers et~al\mbox{.}(2025)]%
        {kommers2025computational}
\bibfield{author}{\bibinfo{person}{Cody Kommers}, \bibinfo{person}{Ruth Ahnert}, \bibinfo{person}{Maria Antoniak}, \bibinfo{person}{Emmanouil Benetos}, \bibinfo{person}{Steve Benford}, \bibinfo{person}{Mercedes Bunz}, \bibinfo{person}{Baptiste Caramiaux}, \bibinfo{person}{Shauna Concannon}, \bibinfo{person}{Martin Disley}, \bibinfo{person}{James Dobson}, {et~al\mbox{.}}} \bibinfo{year}{2025}\natexlab{}.
\newblock \showarticletitle{Computational Hermeneutics: Evaluating Generative {AI} as a Cultural Technology}.
\newblock  (\bibinfo{year}{2025}).
\newblock


\bibitem[Leahu(2016)]%
        {leahu2016ontological}
\bibfield{author}{\bibinfo{person}{Lucian Leahu}.} \bibinfo{year}{2016}\natexlab{}.
\newblock \showarticletitle{Ontological surprises: A relational perspective on machine learning}. In \bibinfo{booktitle}{\emph{Proceedings of the 2016 ACM conference on designing interactive systems}}. \bibinfo{pages}{182--186}.
\newblock


\bibitem[Lewis(2025)]%
        {lewis2025art}
\bibfield{author}{\bibinfo{person}{Makayla Lewis}.} \bibinfo{year}{2025}\natexlab{}.
\newblock \showarticletitle{Art, Identity, and AI: Navigating Authenticity in Creative Practice}. In \bibinfo{booktitle}{\emph{Proceedings of the 2025 Conference on Creativity and Cognition}}. \bibinfo{pages}{916--930}.
\newblock


\bibitem[Manovich(2020)]%
        {manovich2020cultural}
\bibfield{author}{\bibinfo{person}{Lev Manovich}.} \bibinfo{year}{2020}\natexlab{}.
\newblock \bibinfo{booktitle}{\emph{Cultural analytics}}.
\newblock \bibinfo{publisher}{Mit Press}.
\newblock


\bibitem[Michel et~al\mbox{.}(2011)]%
        {michel2011quantitative}
\bibfield{author}{\bibinfo{person}{Jean-Baptiste Michel}, \bibinfo{person}{Yuan~Kui Shen}, \bibinfo{person}{Aviva~Presser Aiden}, \bibinfo{person}{Adrian Veres}, \bibinfo{person}{Matthew~K Gray}, \bibinfo{person}{Google~Books Team}, \bibinfo{person}{Joseph~P Pickett}, \bibinfo{person}{Dale Hoiberg}, \bibinfo{person}{Dan Clancy}, \bibinfo{person}{Peter Norvig}, {et~al\mbox{.}}} \bibinfo{year}{2011}\natexlab{}.
\newblock \showarticletitle{Quantitative analysis of culture using millions of digitized books}.
\newblock \bibinfo{journal}{\emph{science}} \bibinfo{volume}{331}, \bibinfo{number}{6014} (\bibinfo{year}{2011}), \bibinfo{pages}{176--182}.
\newblock


\bibitem[Moretti(2013)]%
        {moretti2013distant}
\bibfield{author}{\bibinfo{person}{Franco Moretti}.} \bibinfo{year}{2013}\natexlab{}.
\newblock \bibinfo{booktitle}{\emph{Distant reading}}.
\newblock \bibinfo{publisher}{Verso Books}.
\newblock


\bibitem[Odom et~al\mbox{.}(2019)]%
        {odom2019unpacking}
\bibfield{author}{\bibinfo{person}{William Odom}, \bibinfo{person}{Ishac Bertran}, \bibinfo{person}{Garnet Hertz}, \bibinfo{person}{Henry Lin}, \bibinfo{person}{Amy Yo~Sue Chen}, \bibinfo{person}{Matt Harkness}, {and} \bibinfo{person}{Ron Wakkary}.} \bibinfo{year}{2019}\natexlab{}.
\newblock \showarticletitle{Unpacking the thinking and making behind a slow technology research product with slow game}. In \bibinfo{booktitle}{\emph{Proceedings of the 2019 Conference on Creativity and Cognition}}. \bibinfo{pages}{15--28}.
\newblock


\bibitem[Odom and Duel(2018)]%
        {odom2018design}
\bibfield{author}{\bibinfo{person}{William Odom} {and} \bibinfo{person}{Tijs Duel}.} \bibinfo{year}{2018}\natexlab{}.
\newblock \showarticletitle{On the design of OLO Radio: Investigating metadata as a design material}. In \bibinfo{booktitle}{\emph{Proceedings of the 2018 CHI Conference on Human Factors in Computing Systems}}. \bibinfo{pages}{1--9}.
\newblock


\bibitem[Odom et~al\mbox{.}(2020)]%
        {odom2020exploring}
\bibfield{author}{\bibinfo{person}{William Odom}, \bibinfo{person}{MinYoung Yoo}, \bibinfo{person}{Henry~WJ Lin}, \bibinfo{person}{Tijs Duel}, \bibinfo{person}{Tal Amram}, {and} \bibinfo{person}{Amy Yo~Sue Chen}.} \bibinfo{year}{2020}\natexlab{}.
\newblock \bibinfo{title}{Exploring the Reflective Potentialities of Personal Data with Different Temporal Modalities: A Field Study of Olo Radio.}
\newblock \bibinfo{numpages}{283--295}~pages.
\newblock


\bibitem[Offert and Bell(2020)]%
        {offert2020generative}
\bibfield{author}{\bibinfo{person}{Fabian Offert} {and} \bibinfo{person}{Peter Bell}.} \bibinfo{year}{2020}\natexlab{}.
\newblock \showarticletitle{Generative Digital Humanities.}. In \bibinfo{booktitle}{\emph{CHR}}. \bibinfo{pages}{202--212}.
\newblock


\bibitem[Parikka(2013)]%
        {parikka2013media}
\bibfield{author}{\bibinfo{person}{Jussi Parikka}.} \bibinfo{year}{2013}\natexlab{}.
\newblock \bibinfo{booktitle}{\emph{What is media archaeology?}}
\newblock \bibinfo{publisher}{John Wiley \& Sons}.
\newblock


\bibitem[Perec(1969)]%
        {perec1969disparition}
\bibfield{author}{\bibinfo{person}{Georges Perec}.} \bibinfo{year}{1969}\natexlab{}.
\newblock \bibinfo{booktitle}{\emph{La Disparition}}.
\newblock \bibinfo{publisher}{Deno{\"e}l}.
\newblock


\bibitem[Risam(2018)]%
        {risam2018new}
\bibfield{author}{\bibinfo{person}{Roopika Risam}.} \bibinfo{year}{2018}\natexlab{}.
\newblock \bibinfo{booktitle}{\emph{New digital worlds: Postcolonial digital humanities in theory, praxis, and pedagogy}}.
\newblock \bibinfo{publisher}{Northwestern University Press}.
\newblock


\bibitem[Suh et~al\mbox{.}(2023)]%
        {suh2023sensecape}
\bibfield{author}{\bibinfo{person}{Sangho Suh}, \bibinfo{person}{Bryan Min}, \bibinfo{person}{Srishti Palani}, {and} \bibinfo{person}{Haijun Xia}.} \bibinfo{year}{2023}\natexlab{}.
\newblock \showarticletitle{Sensecape: Enabling multilevel exploration and sensemaking with large language models}. In \bibinfo{booktitle}{\emph{Proceedings of the 36th annual ACM symposium on user interface software and technology}}. \bibinfo{pages}{1--18}.
\newblock


\bibitem[Suzgun et~al\mbox{.}(2025)]%
        {suzgun2025language}
\bibfield{author}{\bibinfo{person}{Mirac Suzgun}, \bibinfo{person}{Tayfun Gur}, \bibinfo{person}{Federico Bianchi}, \bibinfo{person}{Daniel~E Ho}, \bibinfo{person}{Thomas Icard}, \bibinfo{person}{Dan Jurafsky}, {and} \bibinfo{person}{James Zou}.} \bibinfo{year}{2025}\natexlab{}.
\newblock \showarticletitle{Language models cannot reliably distinguish belief from knowledge and fact}.
\newblock \bibinfo{journal}{\emph{Nature Machine Intelligence}} (\bibinfo{year}{2025}), \bibinfo{pages}{1--11}.
\newblock


\bibitem[Thompson(1967)]%
        {thompson1967time}
\bibfield{author}{\bibinfo{person}{EP Thompson}.} \bibinfo{year}{1967}\natexlab{}.
\newblock \showarticletitle{Time, Work-Discipline, and Industrial Capitalism}.
\newblock \bibinfo{journal}{\emph{Past \& Present}} \bibinfo{number}{38} (\bibinfo{year}{1967}), \bibinfo{pages}{56--97}.
\newblock


\bibitem[Underwood(2019)]%
        {underwood2019distant}
\bibfield{author}{\bibinfo{person}{Ted Underwood}.} \bibinfo{year}{2019}\natexlab{}.
\newblock \bibinfo{booktitle}{\emph{Distant horizons: digital evidence and literary change}}.
\newblock \bibinfo{publisher}{University of Chicago Press}.
\newblock


\bibitem[Underwood et~al\mbox{.}(2025)]%
        {underwood2025can}
\bibfield{author}{\bibinfo{person}{Ted Underwood}, \bibinfo{person}{Laura~K Nelson}, {and} \bibinfo{person}{Matthew Wilkens}.} \bibinfo{year}{2025}\natexlab{}.
\newblock \showarticletitle{Can Language Models Represent the Past without Anachronism?}
\newblock \bibinfo{journal}{\emph{arXiv preprint arXiv:2505.00030}} (\bibinfo{year}{2025}).
\newblock


\bibitem[{U.S. Environmental Protection Agency}(2025)]%
        {epa2025egrid}
\bibfield{author}{\bibinfo{person}{{U.S. Environmental Protection Agency}}.} \bibinfo{year}{2025}\natexlab{}.
\newblock \bibinfo{title}{Emissions \& Generation Resource Integrated Database ({eGRID}), Year 2023 Data}.
\newblock \bibinfo{howpublished}{\url{https://www.epa.gov/egrid}}.
\newblock
\newblock
\shownote{Released January 2025}.


\bibitem[Weick and Weick(1995)]%
        {weick1995sensemaking}
\bibfield{author}{\bibinfo{person}{Karl~E Weick} {and} \bibinfo{person}{Karl~E Weick}.} \bibinfo{year}{1995}\natexlab{}.
\newblock \bibinfo{booktitle}{\emph{Sensemaking in organizations}}. Vol.~\bibinfo{volume}{3}.
\newblock \bibinfo{publisher}{Sage publications Thousand Oaks, CA}.
\newblock


\bibitem[Zhou et~al\mbox{.}(2025)]%
        {zhou2025thoughtful}
\bibfield{author}{\bibinfo{person}{David Zhou}, \bibinfo{person}{John~R Gallagher}, {and} \bibinfo{person}{Sarah Sterman}.} \bibinfo{year}{2025}\natexlab{}.
\newblock \showarticletitle{Thoughtful, Confused, or Untrustworthy: How Text Presentation Influences Perceptions of AI Writing Tools}. In \bibinfo{booktitle}{\emph{Proceedings of the 2025 Conference on Creativity and Cognition}}. \bibinfo{pages}{573--589}.
\newblock


\end{thebibliography}
\end{document}